# Protection d'un système d'information par une intelligence artificielle : une approche en trois phases basée sur l'analyse des comportements pour détecter un scénario hostile

## Airbus Defence and Space — CyberSecurity, France

Jean-Philippe Fauvelle : www.linkedin.com/in/jpfauvelle/
Alexandre Dey : www.linkedin.com/in/alexandre-dey/
Sylvain Navers : www.linkedin.com/in/sylvainnavers/

**Abstract.** L'analyse des comportements des personnes et des entités (UEBA, en anglais) est un domaine de l'intelligence artificielle qui permet de détecter des actions hostiles (ex. : attaques, fraudes, influence, empoisonnement) grâce au caractère inhabituel des évènements observés, par apposition à un fonctionnement basé sur des signatures. Un procédé UEBA comprend habituellement deux phases, d'apprentissage et d'inférence. Les systèmes de détection d'intrusion (IDS, en anglais) du marché souffrent encore de biais notamment d'une sur-simplification des problématiques, d'une sous-exploitation du potentiel de l'IA, d'une prise en compte insuffisante de la temporalité des évènements, et d'une gestion perfectible du cycle de la mémoire des comportements. En outre, alors qu'une alerte générée par un IDS à base de signatures peut se référer à l'identifiant de la signature sur laquelle se fonde la détection, les IDS du domaine UEBA produisent des résultats, souvent associés à un score, dont le caractère explicable est moins évident. Notre approche, non supervisée, consiste à enrichir ce procédé en lui adjoignant une troisième phase permettant de corréler des évènements (incongruités, signaux faibles) présumés liés entre eux, avec pour bénéfice une réduction des faux positifs et négatifs. Nous cherchons également à éviter un biais dit « de la grenouille ébouillantée » inhérent à l'apprentissage continu. Nos premiers résultats sont intéressants et revêtent un caractère explicable, autant sur des données synthétiques que réelles.

## 1 Introduction

Depuis l'avènement d'Internet et des plateformes en ligne telles que Netflix ou Amazon, des avancées technologiques ont été réalisées dans le but de comprendre, catégoriser et prédire automatiquement les comportements des utilisateurs et des entités, posant les fondations du domaine *User and Entity Behavior Analytics* (UEBA).

Depuis la publication par Gartner en 2014 du *Market Guide for User Behavior Analytics* [1], le périmètre d'application du domaine UEBA le plus connu est la cyber-sécurité, notamment par son aptitude à repérer les fuites de données et à lutter contre une forme croissante de menace intrusive nommée *Insider Threat* réalisée depuis l'intérieur du SI (ex. : vol de mots de passe, *phishing*, utilisateurs malicieux).

Sur le plan technique, des algorithmes de *machine learning* pour la détection d'anomalies (ex. : *Local Outlier Factor* [2], *Isolation Forest* [3], réseaux de neurones auto-encodeurs [4]) sont utilisés afin de modéliser les comportements normaux pour détecter lorsque ces derniers dévient de leurs habitudes. Les modèles sont construits de manière non-supervisée (sans étiquetage préalable des données) à partir des informations relevées par un ensemble de capteurs (ex. : sonde réseau, log de connections).

Il existe à ce jour plusieurs solutions de cyber-sécurité intégrant de l'UEBA, dont les principales sont Splunk [5], IBM QRadar [6], Exabeam [7] et LogRythm [8].

Cependant, ces solutions possèdent des limites, à la fois sur les algorithmes utilisés, mais aussi et surtout sur leurs implémentations et utilisations.

Par exemple, dans le cas de OneClass-SVM [9], le détecteur d'anomalies intégré à Splunk, l'algorithme est lourd à entraîner, avec une complexité évoluant quadratiquement en fonction du nombre d'éléments d'entraînement, et il est de surcroit peu adapté à traiter des problèmes non-linéaires.

Ces contraintes algorithmiques se répercutent sur la manière d'intégrer ces technologies aux SIEM (*Security Information and Event Management*).

Une complexité d'entraînement trop forte obligera à limiter le volume de données de référence, réduisant la qualité du modèle appris, et pouvant éventuellement faciliter la pollution de la base d'apprentissage. De façon similaire, une faiblesse algorithmique imposera de travailler les données d'entrée de manière à contourner ces difficultés, tâche ardue pour un individu sans formation avancée en data science, et qui créera de surcroît des angles morts exploitables.

Aux limites susmentionnées des solutions existantes s'ajoutent, d'une part un taux de faux positifs souvent élevé et d'autre part une fréquente sur-simplification des problèmes à résoudre, qui caractérisent la problématique majeure de ces solutions.

On peut principalement témoigner de ces biais à quatre niveaux, à commencer par la présence quasi-systématique d'un compteur d'alertes par fenêtre temporelle (i.e. levée d'une alerte en présence, dans cette fenêtre temporelle, d'un nombre d'anomalies supérieur à un seuil prédéfini) inadapté à la détection d'attaques complexes (ciblées, peu bruyantes, sur plusieurs mois). En deuxième lieu, ces systèmes sont conçus en améliorant les différentes composantes des systèmes traditionnels, plutôt qu'en exploitant les nouvelles possibilités offertes par l'IA pour proposer une approche novatrice. En troisième lieu, ils prennent insuffisamment en compte la temporalité des évènements, porteuse d'information. Enfin, leur gestion du cycle de la mémoire des comportements (i.e. apprentissage et oublis progressifs) reste limitée.

Par ailleurs, et quelle que soit la qualité de détection d'un procédé UEBA, les solutions IDS actuelles, produisant des alertes notamment matérialisées par un score insuffisamment étayé, ne sont pas aisément utilisables, pour deux raisons, humaine et informationnelle : on comprend aisément les réticences d'un individu à engager des actions préventives ou correctives portant à conséquence, en se fondant sur une alerte scorée sortant d'une boite noire, à fortiori lorsque la chaîne d'évènements causaux ayant conduit à l'alerte (i.e. caractère explicable) n'est pas clairement identifiée.

## 2     Principe général

Un procédé UEBA comprend généralement une phase d'apprentissage et une d'inférence qui mettent en œuvre une capacité à apprendre les nouveaux comportements : un comportement qui perdure devient progressivement la norme (i.e. habituel, normal) tandis qu'un comportement abandonné devient incongru (i.e. inhabituel, anormal). Il en découle les principes de superposition des états (normalité vs incongruité), de progressivité dans leurs transitions, de capacité à assimiler et à oublier.

Bien que les modèles soient construits de manière non-supervisée, la conception initiale de l'IA nécessite une étape d'évaluation par apprentissage supervisé. Or l'apprentissage des habitudes porte sur toutes les données et peut s'étendre sur de larges périodes. Labéliser manuellement un tel ensemble de données est une gageure. Pour ce motif et compte tenu du caractère sensible des données inhérentes au domaine de la cyber-sécurité, se procurer des données massives, cohérentes, labélisées, et prêtes à l'emploi, constitue une difficulté non insurmontable mais majeure.

La phase d'apprentissage alimente un référentiel des comportements habituels dit « référentiel comportemental », en évolution permanente, tandis que la phase d'inférence, pour chaque donnée collectée, génère son score d'incongruité en fonction de sa déviation relativement au comportement habituel connu du référentiel.

Cette inférence basée sur ledit référentiel mouvant, permet d'assimiler les nouveaux comportements. Cette qualité possède un biais dit « de la grenouille ébouillantée[1] » [10] inhérent aux besoins antinomiques, d'une part d'assimiler rapidement les changements de comportements afin de s'y adapter, et d'autre part de les assimiler lentement pour rester capable de détecter le différentiel caractérisant tout changement.

L'effet conjoint des phases d'apprentissage et d'inférence automatiques permet simultanément, d'une part d'apprendre en continue les comportements c'est-à-dire de s'adapter à leur évolution, et d'autre part d'attribuer à chaque comportement observé à l'instant, un score d'incongruité qui quantifie la déviation entre ledit comportement et les comportements appris antérieurement et connus à cet instant.

Il convient de souligner que cet apprentissage est agnostique vis-à-vis des notions de bien et de mal, puisque il se borne à détecter des transitions de comportements. À l'extrême, un comportement hostile qui perdure devient progressivement la norme, ceci étant un défaut intrinsèque à l'UEBA.

---

[1]   Une grenouille plongée dans l'eau bouillante s'en échappe, mais ne réagit pas dans une eau froide progressivement chauffée. Cette fable au fondement scientifique partiel, est cependant utilisée par des études sérieuses pour illustrer que l'accoutumance progressive à un changement continu assez lent peut empêcher la détection d'une situation pourtant nocive [10].

## 3 Approche proposée

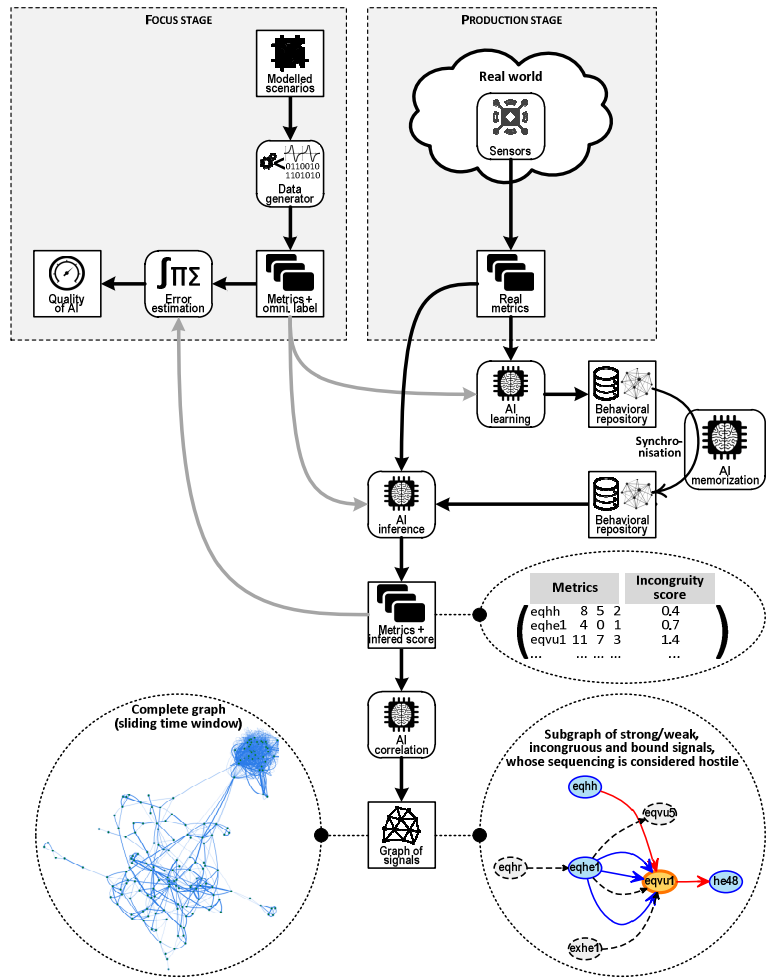

**Fig. 1.** Synoptique de l'approche proposée.

### 3.1 Problématique des données

**Métriques**

Dans notre approche générique et volontairement simplifiée, les métriques alimentant l'apprentissage et l'inférence sont issues de capteurs de tous domaines y compris non cyber. Elles comprennent les identifiants des équipements véhiculant les flux (source, destination), les informations inhérentes à la messagerie (émetteur apparent, destinataire, pièce jointe), les protocoles et ports de communication, les horodatages, ainsi qu'une information issue du monde du renseignement qui permet d'établir une rela-tion indirecte probable entre l'équipement utilisé par le hacker et un milieu connu pour avoir une activité sensible.

**Modélisation et génération des données synthétiques**

Afin de remédier à la carence en données cohérentes et labélisées, nous avons conçu un générateur de données synthétiques, basé sur des règles programmables à partir des connaissances opérationnelles des comportements normaux et anormaux (signaux forts et faibles, bruits, pollutions). Ce générateur produit en masse des métriques multiformes et cohérentes qui reflètent la variabilité et les interactions entre les personnes et entités à l'origine des évènements. En outre, chaque règle programmée ayant connaissance du caractère hostile ou non des métriques qu'elle produit, permet au générateur d'insérer un label dit omniscient dans les métriques, indispensable à la conception initiale de l'IA. Une fois l'IA conçue, les données réelles prennent le pas sur les données synthétiques et le procédé demeure non supervisé.

**Enrichissement des données**

Nous améliorons la qualité de l'apprentissage notamment en le focalisant sur l'ordre de grandeur des métriques numériques et en ajoutant des métriques obtenues par agrégations (ex. : dénombrements de requêtes, sommes de volumes) de métriques existantes sur des fenêtres temporelles glissantes, ce qui permet de prendre en compte les fluctuations relatives des évènements.

### 3.2  Problématique inhérentes à l'IA

**Évaluation automatique de la qualité du résultat**

L'insertion d'un label omniscient dans les métriques par notre générateur facilite grandement la conception de l'IA, étant rappelé que ce label omniscient est utilisé exclusivement pour la conception de l'IA. En effet, la comparaison, pour chaque métrique, du label en entrée de l'IA et du score inféré en sortie, permet de quantifier automatiquement l'écart entre l'attendu et l'inféré, ledit écart constituant une estimation de la qualité du procédé d'IA (**Fig. 1**).

Dans notre approche actuelle, qui pourra évoluer, cette estimation est basée sur l'erreur quadratique moyenne positive et négative, calculée distinctement pour les métriques correspondant à des comportements hostiles et celles correspondant à des comportements normaux, conformément au label omniscient.

**Phases d'apprentissage et d'inférence**

Après enrichissement, les métriques sont converties au format numérique. En considérant ces variables comme des coordonnées, chaque événement peut ainsi être associé à un point dans un espace multidimensionnel, les points qui correspondent à des comportements normaux tendant à se regrouper en paquets nommés clusters, tandis que ceux correspondant à des anomalies restent à l'écart desdits clusters.

L'algorithme mis en œuvre dans notre expérimentation sur données synthétiques se nomme *Isolation Forest* [3], figure parmi les plus précis de l'état de l'art académique, et reste rapide à entraîner. Il repose sur le principe qu'il est plus facile d'isoler un élément incongru des autres éléments plutôt qu'un élément au centre d'un cluster.

Au cours de l'entraînement, cet algorithme effectue des coupes aléatoires successives en sélectionnant à chaque fois une dimension au hasard (ex. : élimination des éléments pour lesquels la composante sélectionnée est inférieure à la valeur aléatoire) jusqu'à ce qu'il ne subsiste qu'un point dans la coupe ou que le nombre de coupes maximal soit atteint. Chaque coupe successive ajoute une branche à un arbre de décision. Afin d'améliorer la précision et d'éviter de fausser les résultats par une coupe malchanceuse, l'algorithme crée plusieurs arbres rassemblés au sein d'une forêt après élagage des branches inutiles.

Lors de l'inférence, l'algorithme dénombre les branches à parcourir pour atteindre la feuille de l'arbre correspondant au point inféré, le niveau d'incongruité dudit point étant plus élevé lorsque la longueur du chemin parcouru est plus faible.

**Score d'incongruité**

Le score d'incongruité, ou score d'inférence, produit en sortie de la phase d'inférence et transmis à la phase de corrélation, est un nombre réel positif ou nul, sans unité, qui caractérise le niveau d'incongruité d'un n-uplet de données par rapport aux données précédemment apprises. L'échelle de valeurs du score, et en particulier sa plus grande valeur possible, sont directement liés aux procédés d'apprentissage et d'inférence.

La normalisation du score d'incongruité est une problématique subtile qui joue sur l'efficacité du procédé global. Dans un premier temps, nous avions décidé de normaliser ce score et d'arrondir à zéro ses faibles valeurs, mais cette perte d'information a affaibli l'apport de la phase de corrélation. Nous avons finalement choisi de ne pas normaliser le score et d'enrichir notre phase de corrélation par une fonction auto adaptative qui accepte en entrée un score non normalisé.

Ainsi, l'échelle de valeurs du score n'est plus primordiale, seule comptant le caractère relatif entre eux des scores d'incongruité des données inférées. À titre d'illustration à partir de résultats présentés au §4.2, un n-uplet de données dont le score d'incongruité est égal à 0.6589 est légèrement plus incongru qu'un score moyen à 0.504 et nettement moins incongru qu'un n-uplet scoré à 1.6233.

Cette solution est avantageuse en ce qu'elle découple les phases d'inférence et de corrélation en permettant de changer le procédé d'inférence sans être contraint d'adapter celui de corrélation. Cette facilité sera utilisée (cf §5).

**Phase de corrélation**

En complément des phases d'apprentissage et d'inférence, nous ajoutons une phase dite de corrélation, alimentée par le résultat de l'inférence et produisant des graphes associés aux scores de pertinence. Cette phase de corrélation comprend trois espaces.

Le premier espace traite les données produites par la phase d'inférence, sous la forme de n-uplets de métriques dont chacun est assorti du score d'incongruité non normalisé issu de l'inférence. Un procédé regroupe les n-uplets assimilés à des quasi-jumeaux pour faciliter le passage à l'échelle de la corrélation.

Le second espace traite les n-uplets qui d'une part semblent liés par des dépendances probables et d'autre part sont considérés d'intérêt majeur relatif. Dans notre première version à ce jour, l'existence de certaines variables identiques dans des n-uplets distincts, implique l'existence d'un lien entre les évènements à l'origine des n-uplets. L'intérêt majeur relatif d'un n-uplet est corrélé positivement à son score d'incongruité par l'intermédiaire d'une fonction de pertinence (FP) confidentielle, auto adaptative, à rétroaction temporelle, qui prend en compte l'absence de normalisation dudit score notamment via un effet hystérétique et d'oubli.

Le troisième espace traite les graphes de pertinence majeure, composés de n-uplets quelconques (majeurs ou mineurs c.-à-d. signaux forts ou faibles) reliés entre eux. La pertinence d'un graphe est quantifiée via la fonction FP qui prend en compte de nombreux facteurs (ex. : nombre de n-uplets, échelles temporelles, scores d'incongruité des n-uplets, caractère faible/fort des signaux, caractéristiques topologiques, probabilités, etc.) dont certains s'atténuent ou s'amplifient entre eux.

C'est ainsi qu'un signal faible peut, au sein d'une longue chaîne d'évènements matérialisée par un graphe, signifier peu individuellement mais renforcer significativement le tout. Ceci justifiant l'importance de ne pas sur-simplifier (ex. : filtrer, arrondir) les informations, même faiblement scorées, en sortie de l'inférence.

In fine, ce procédé permet d'ignorer des métriques possédant un score d'incongruité majeur en sortie d'inférence mais ne contribuant pas à un graphe de pertinence majeure, et inversement de tenir compte des métriques peu incongrues (signaux faibles) mais contribuant à un graphe pertinent. À l'extrême, même un signal faible voire nul, peut contribuer à un graphe pertinent du seul fait qu'il permet une jonction entre certains évènements et en amplifie d'autres, et ce, même dans un graphe ne comportant aucun signal fort.

En d'autres termes, notre approche permet d'amplifier les faux négatifs et d'affaiblir les faux positifs selon leur contribution à des chaînes favorablement scorées d'évènements reliés de manière logique et temporelle, y compris sur des fenêtres temporelles potentiellement longues (i.e. des mois).

Ainsi, nous prenons en compte l'information issue de la temporalité des données et nous évitons l'autre biais qui consiste à simplement dénombrer des alertes dans une fenêtre temporelle et qui est à l'origine de faux positifs et faux négatifs.

**De la grenouille ébouillantée à la gestion de la mémoire des comportements**

À ce jour, nous contournons le biais de la grenouille ébouillantée en dupliquant le référentiel comportemental, une instance alimentée par l'apprentissage étant synchronisée périodiquement vers une autre consultée par l'inférence (**Fig. 1**).

Ainsi, l'inférence compare un comportement présent avec une instance légèrement obsolète du référentiel, de sorte que l'écart de comportement mesuré atteigne un seuil détectable, dont l'amplitude dépend d'une part de la vitesse à laquelle évolue ledit comportement et d'autre part de l'obsolescence de l'instance du référentiel c'est-à-dire du délai entre les synchronisations.

Dans cette première approche simpliste, le paramétrage du délai de synchronisation est statique, global, et relève d'un compromis peu satisfaisant. Ce procédé pourra être amélioré en instaurant un délai de synchronisation dynamique et en le différenciant pour chaque individu et équipement identifié.

Afin de mieux exploiter le potentiel de l'intelligence artificielle et nous démarquer des approches UEBA actuelles, et en s'inspirant du fonctionnement cyclique du processus de mémorisation dans le cerveau humain, nous réfléchissons à la conception d'une IA de mémorisation permettant d'optimiser assimilation, renforcement et oubli.

**Caractère explicable du résultat**

Le résultat produit en sortie de notre phase de corrélation prend la forme de graphes d'évènements reliés, aisément interprétables par un individu (**Fig. 3**).

**Passage à l'échelle**

Notre phase d'apprentissage est actuellement limitée à une multi-instanciation sur les types des données, mais nos phases d'inférence et de corrélation possèdent déjà une capacité de mise à l'échelle horizontale qui lève l'obstacle d'un traitement massif.

## 4    Nos premiers résultats

### 4.1    Scénario

Afin de soumettre notre approche à une preuve de concept, nous avons élaboré un scénario de compromission d'informations dans un SI d'entreprise (**Fig. 2**). Ce scénario reste volontairement simple afin de conserver un caractère facilement explicable, la finalité étant de mettre en exergue les apports de la phase de corrélation.

Dans ce scénario (**Fig. 2**), un hacker (HH) effectue un criblage/ciblage (A10, A11) sur un réseau social (EQSN), puis usurpe l'identité d'un individu référent (HR) et adresse (A13) à plusieurs employés (*) un courriel comprenant une pièce jointe malicieuse, laquelle est activée par un employé (HE1) puis effectue un scan (*) de ports (A17) rendant possible l'exploitation d'une vulnérabilité sur un équipement réseau (EQVU1), permettant au hacker de collecter (A19) des documents sensibles sur le SI.

(*) La pièce jointe malicieuse est également adressée à l'employé HE2 mais ce dernier ne l'active pas. Le scan de ports est volontairement effectué au ralenti, sur une période de plusieurs heures, afin d'être peu incongru donc faiblement détectable.

Ce scénario modélisé comprend trois incongruités comportementales susceptibles d'être détectées ($BI_1$, $BI_2$, $BI_3$ en **Fig. 2**). Il se déroule sur deux mois dont deux jours dévolus à l'attaque, dans une entreprise constituée d'une centaine d'employés travaillant en interne et depuis leurs domiciles, sur un SI représentatif (PC internes/externes, messagerie, flux réseau, firewall, routeurs), et comprend des flux internes, externes, mixtes ainsi que du renseignement (source ouverte) en rapport avec le hacker.

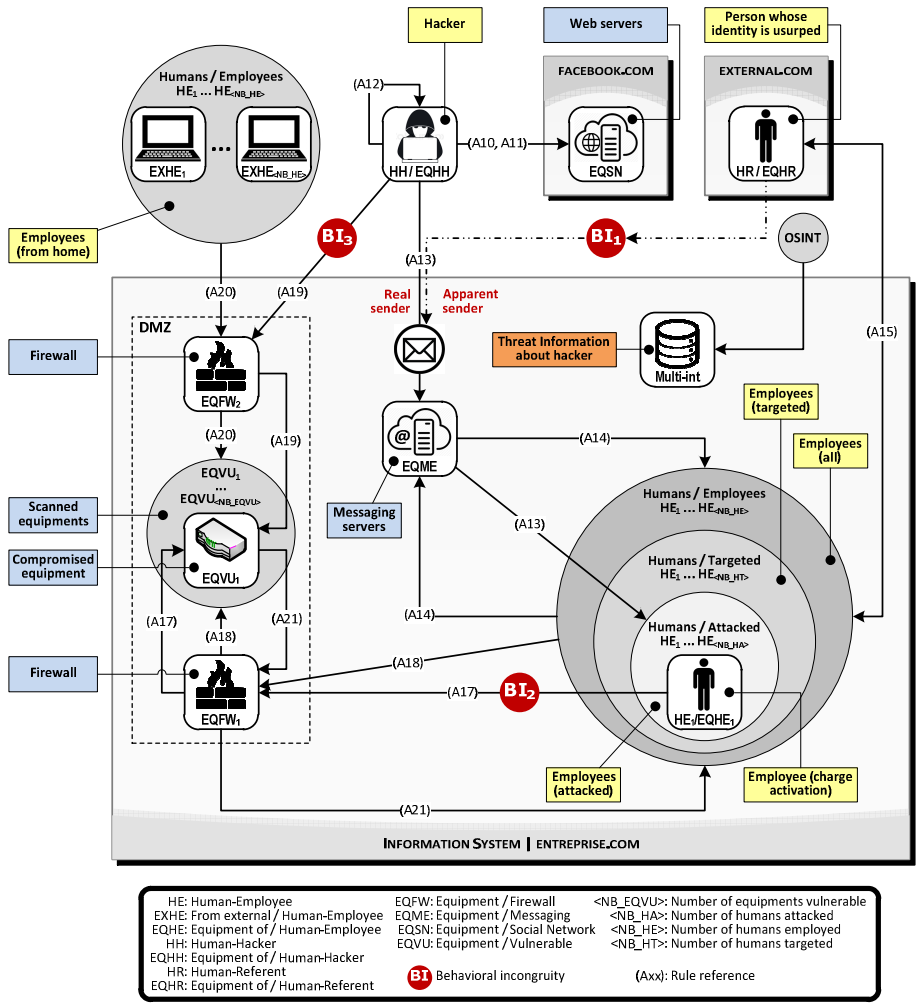

**Fig. 2.** Synoptique du scénario modélisé lors du POC (données synthétiques).

Ce scénario simulé produit environ 500,000 métriques correspondant à des évènements cohérents, des incongruités, des signaux faibles, ainsi qu'un fort volume de pollution noyant tous les évènements hostiles dans ceux des comportements normaux.

### 4.2 Résultat

Le résultat en sortie d'inférence est un graphe d'intérêt majeur (**Fig. 3**, **Tableau 1**).

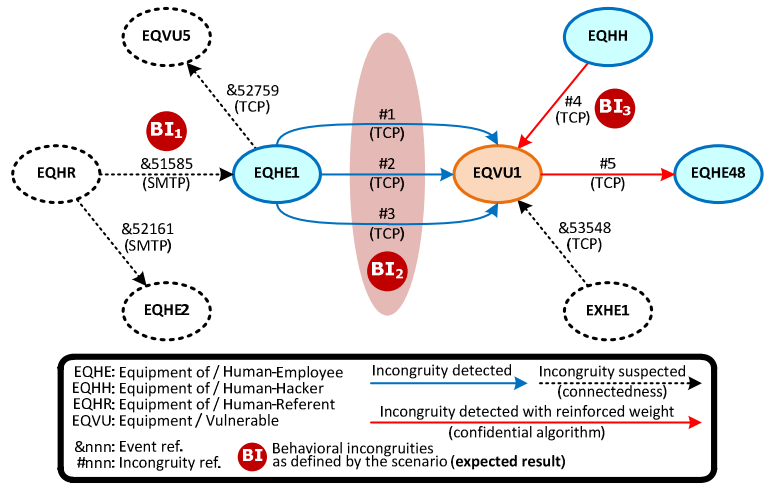

**Fig. 3.** Graphe d'intérêt majeur produit par l'IA de corrélation (données synthétiques). Les évènements sont détaillés dans le **Tableau 1**. Les mentions BI$_n$ sont ajoutées manuellement pour faciliter les recoupements avec le scénario en **Fig. 2**. Une pastille « EQ... » symbolise un équipement (cf. nomenclature de la figure). Une flèche symbolise un évènement référencé « &nnn » qui correspond à un flux, lequel est un n-uplet de données. Une flèche bleue ou rouge correspond à un évènement incongru référencé « #nnn » (numérotation chronologique) c.-à-d. à un n-uplet d'intérêt majeur ou mineur selon son score. Une flèche rouge, par opposition à une bleue, correspond à une incongruité <u>renforcée</u>. Une flèche noire en pointillés correspond à un évènement jugé simplement <u>suspect</u> en raison de sa connexité et contribuant à ce titre au graphe d'intérêt majeur.

**Tableau 1.** Données associées aux évènements de la **Fig. 3**.

| Incon-gruité | Détection | Réf. évène-ment | Horoda-tage | Score | Centile | Équipe-ment Source / Dest. | Proto-cole | Meta |
|---|---|---|---|---|---|---|---|---|
| **BI$_1$** | Suspect | &51585 | 2017-05-02 09:08:19 | 0.6589 | 0.9071 | EQHR/ EQHE1 | SMTP | has_attach : true attach_size : 42779 mail_from : HR@external.com mail_to : HE1@entreprise.com |
| | Suspect | &52161 | 2017-05-02 11:35:13 | 0.6589 | 0.9072 | EQHR/ EQHE2 | SMTP | has_attach : true attach_size : 45798 mail_from : HR@external.com mail_to : HE2@entreprise.com |
| **BI$_2$** | Incongru #1 | &52282 | 2017-05-02 12:25:46 | 0.5852 | 0.9292 | EQHE1/ EQVU1 | TCP | port : 443 |
| | Suspect | &52759 | 2017-05-02 16:00:42 | 0.5775 | 0.9014 | EQHE1/ EQVU5 | TCP | port : 19072 |
| **BI$_2$** | Incongru #2 | &52778 | 2017-05-02 16:04:20 | 0.5852 | 0.9296 | EQHE1/ EQVU1 | TCP | port : 4170 |
| **BI$_2$** | Incongru | &52899 | 2017-05-02 | 0.6014 | 0.9592 | EQHE1/ | TCP | port : 41854 |

|     |     |     | #3 |     | 16:22:49 |     |     | EQVU1 |     |     |
| --- | --- | --- | --- | --- | --- | --- | --- | --- | --- | --- |
| $BI_3$ |     |     | Incongru #4 | &53530 | 2017-05-02 22:20:28 | 1.6233 | 1.0000 | EQHH/ EQVU1 | TCP | port : 46761 |
|     |     |     | Suspect | &53548 | 2017-05-02 23:09:09 | 0.4881 | 0.9842 | EXHE1/ EQVU1 | TCP | port : 22 |
|     |     |     | Incongru #5 | &53559 | 2017-05-02 23:34:27 | 0.6884 | 0.9984 | EQVU1/ EQHE48 | TCP | port : 2279 |

Les mentions $BI_n$ correspondent aux incongruités générées par le scénario, c'est-à-dire aux évènements dont la détection est attendue.

Une première analyse de ce résultat (**Fig. 3**, **Tableau 1**) montre que le scénario d'origine (**Fig. 2**) est bien détecté :

— Le graphe d'intérêt majeur fait bien apparaitre l'incongruité $BI_1$ du scénario d'origine, correspondant à l'envoi de la pièce jointe malicieuse (&51585) semblant provenir de l'individu référent (EQHR) dont l'identité est usurpée. Cependant, cet évènement n'est pas estimé incongru mais simplement suspect (procédé confidentiel), et à ce titre il contribue tout de même au résultat.
— L'incongruité $BI_2$ du scénario d'origine, correspondant au scan de ports vers l'équipement vulnérable (EQVU1), est bien détectée (#1, #2, #3).
— L'incongruité $BI_3$ du scénario d'origine, correspondant l'exploitation de la vulnérabilité par le hacker (EQHH), est bien détectée (#4).

Les évènements correspondant aux actions hostiles $BI_1$ (envoi de la pièce jointe malicieuse) et $BI_2$ (scan de ports peu incongru effectué au ralenti) demeurent des signaux relativement modérés (score d'incongruité égal à 0.6589 pour $BI_1$ et 0.5852 à 0.6014 pour $BI_2$, à comparer au score d'incongruité moyen sur l'ensemble des données égal à 0.504 avec un écart-type relatif de 11%) à l'exception de l'exploitation de la vulnérabilité (A19 en **Fig. 2** ; #4 en **Fig. 3**) pour laquelle le score d'incongruité est égal à 1.6233 (colonne « score » du **Tableau 1**).

Ainsi, bien que seul $BI_3$ constitue un signal fort, les trois actions hostiles associées à $BI_1$, $BI_2$ et $BI_3$ sont détectées en tant qu'évènements incongrus ($BI_2$ et $BI_3$) ou à minima suspects ($BI_1$) et toutes contribuent à ce graphe d'intérêt majeur.

Une seconde analyse du résultat met en lumière deux informations intéressantes :

— D'une part, le graphe d'intérêt majeur montre que le flux correspondant à l'envoi de la pièce jointe malicieuse (&51585) activée par l'employé (EQHE1) possède un jumeau quasi concomitant (&52161) vers le poste de travail (EQHE2) avec une pièce jointe de taille similaire, ce qui laisse présumer la présence de la charge malicieuse sur ledit poste. Cet évènement estimé suspect est un résultat intéressant qui, dans le cadre d'une exploitation par un SOC, inciterait à vérifier le poste de travail (EQHE2) sur lequel la charge non activée se trouve peut-être encore.
— D'autre part, le résultat montre la détection d'un évènement incongru (#5) correspondant à l'action de collecte des informations sensibles par le hacker. En effet, cette collecte comprend deux flux, l'un entre l'équipement (EQHH) du hacker et l'équipement réseau (EQVU1) ($BI_3$ et A19 en **Fig. 2**), et un second entre (EQVU1) et le poste de travail (EQHE48) de l'employé (HE48) sur lequel les informations sont effectivement collectées (A21 en **Fig. 2** ; #5 en **Fig. 3**). Nous n'avions pas identifié ce dernier comme une incongruité, à tort. À juste titre, notre procédé a évalué que cet évènement constitue bien une incongruité, et l'a intégré de manière contributive au graphe de pertinence majeur caractérisant un scénario hostile.

L'association entre inférence et corrélation reconstitue une chaîne d'évènements liés, comprenant certains signaux faibles qui pourraient demeurer invisibles s'ils n'étaient considérés qu'individuellement.

Ainsi, notre approche permet de détecter des chaînes comprenant peu de signaux forts (un seul dans l'exemple présenté, deux en comptant le signal que nous avions oublié) voire aucun, grâce à l'exploitation de l'information issue des liens probables et de la temporalité.

## 5    Extension aux données réelles

### 5.1    Approche proposée

Afin de confirmer nos résultats obtenus à partir de données synthétiques issues de la simulation d'une activité sur un SI, nous avons poursuivi notre expérimentation à partir de données réelles.

Souhaitant par la même occasion illustrer que nos travaux peuvent s'appliquer à de nombreux domaines, les données réelles sont cette fois-ci inhérentes à l'activité d'un utilisateur sur un poste de travail, équipé d'un système d'exploitation Linux et d'applications.

Notre expérimentation à partir des données réelles comporte toujours trois phases, d'apprentissage, d'inférence et de corrélation. Le différenciant majeur réside dans l'utilisation, pour les phases d'apprentissage et d'inférence, d'un procédé d'apprentissage profond (*deep learning*, en anglais) possédant une capacité de mise à l'échelle horizontale, que nous avons intégré à la suite logicielle Elastic Stack [17].

### 5.2    Données

Les informations collectées sont obtenues via les fonctions d'audit du système d'exploitation. Elles comprennent notamment les actions non autorisées, les appels aux fonctions et commandes permettant de modifier le noyau ou ses modules, les actions suspectes (ex. : nmap, wget, tcpdump), l'accès à des fichiers en zones surveillées (ex. : configurations, binaires, fichiers temporaires), les commandes exécutées, et les invocations de certaines primitives potentiellement dangereuses. Les informations d'identification (ex. : utilisateur, groupe) et de contexte (ex. : chemin d'accès, horodatage, processus parent) sont également collectées.

Ces données sont porteuses de l'information permettant la détection, d'une part des actions inhabituelles (ex. : modification du noyau, d'un fichier de configuration, suppression de fichiers potentiellement témoins d'une intrusion), d'autre part des actions effectuées par un utilisateur inhabituel (ex. : utilisateur non privilégié exécutant ou essayant des commandes privilégiées) et enfin des actions effectuées par une application inhabituelle (ex. : tentative de lecture d'un fichier sensible via le service internet).

### 5.3 Scénario d'entrainement

Un utilisateur réel utilise un poste de travail pendant une semaine, effectuant majoritairement des tâches bureautiques (ex. : traitement de texte, navigation Internet, messagerie) et occasionnellement des exécutions de commandes et de scripts. Cette activité génère environ 200,000 évènements par jour, dont 90% d'appels à des primitives du système d'exploitation.

### 5.4 Scénario d'attaque

Le scénario d'attaque comprend les actions suivantes exécutées par l'utilisateur, dans cet ordre :

— L'utilisateur est dupé et exécute un script malveillant ;
— Ce script télécharge un exploit sur internet via le programme *wget* ;
— Cet exploit est compilé avec le programme *gcc* ;
— Le code compilé résultant est exécuté et cherche à élever ses privilèges en exploitant une vulnérabilité du noyau du système d'exploitation.

### 5.5 Métriques

Les différentes données collectées peuvent prendre un nombre fini de valeurs, lesquelles sont dénuées de relation d'ordre entre elles. Ces variables dites catégorielles doivent nécessairement être converties en variables numériques afin de pouvoir être traitées par les algorithmes de détection. Cette conversion est effectuée par calcul des probabilités d'observer certaines valeurs ayant observé d'autres valeurs. Ce calcul statistique est effectué sous deux angles d'observation :

— La probabilité d'observer un couple de valeurs (nom de l'action, ensemble des identifiants) ayant observé le chemin du programme effectuant l'action ;
— La probabilité d'observer un couple de valeurs (chemin du programme effectuant l'action, ensemble des identifiants) ayant observé le nom de l'action.

Notre procédé de conversion garantit qu'une valeur catégorielle plus fréquemment observée est convertie en une valeur numérique plus forte.

### 5.6 Algorithme de détection d'anomalie par apprentissage profond

L'algorithme *Isolation Forest*, utilisé précédemment dans le cadre de nos travaux menés à partir de données synthétiques issues de la simulation d'une activité sur un SI, ne permet pas l'entraînement parallèle d'un modèle. Compte tenu de cette limitation qui pénalise l'utilisation d'un jeu de données volumineux, nous avons décidé de recourir à un autre procédé, basé sur des réseaux de neurones profonds, lesquels sont entraînables de manière massivement parallèle sur des unités de calcul graphique de type GPU (ex. : TensorFlow [18], MXNET [11]).

Les réseaux de neurones sont des graphes à arêtes pondérées dont les nœuds sont appelés neurones. Ils comprennent une couche d'entrée, une couche de sortie, et d'éventuelles couches intermédiaires dites cachées. L'entraînement d'un réseau de neurone consiste à faire varier les poids des arêtes de façon à ce que pour chaque valeur d'entrée donnée, la valeur de sortie calculée se rapproche au maximum de la valeur attendue. Ils sont principalement utilisés en mode supervisé pour effectuer une classification (i.e. déterminer la classe à laquelle appartient le vecteur d'entrée) ou une régression (calculer une valeur en fonction des données d'entrée).

Il existe cependant un type non-supervisé de réseau de neurones, nommé auto-encodeur, dont le principe consiste à apprendre à répliquer le vecteur d'entrée en faisant transiter ce dernier via une couche dite de compression afin de reconnaître les corrélations qui existent entre toutes les variables de la couche d'entrée.

Dans le cas de la détection d'anomalies, la capacité de dé-bruitage des auto-encodeurs a récemment été mise à profit [12] [13] en s'appuyant sur le fait qu'une donnée bruitée (anomalie) sera mal répliquée par l'auto-encodeur. En mesurant l'écart entre la valeur d'entrée et la valeur de sortie, il devient possible déterminer si une donnée est normale (écart faible) ou non (écart fort).

Nous avons adopté ce procédé de réseau de neurones auto-encodeur qui répond à notre besoin. Afin de garantir la capacité de généralisation ainsi que la résistance des modèles aux attaques de type *adversarial* (i.e. dont la finalité est de duper un procédé de détection), nous employons diverses techniques de régularisation comme le dropout [14], l'insertion de bruit à l'entraînement [15], ou encore l'arrêt anticipé de l'entraînement [16].

### 5.7 Score d'incongruité

Comme décrit précédemment, détecter les anomalies grâce à un auto-encodeur nécessite de mesurer l'écart entre les données entrées et les données répliquées. Le score d'incongruité résulte de la mesure de cette erreur de réplication.

Cependant, il existe différentes méthodes pour mesurer ces erreurs de réplication (ex. : distance Euclidienne, distance de Manhattan, taux d'accroissement) qui peuvent apporter des informations complémentaires (ex. : la distance euclidienne est plus pertinente que la distance de Manhattan lorsque une seule des variables est mal répliquée, par opposition au cas où toutes les variables sont faiblement anormales). Pour établir le score d'incongruité, notre algorithme mesure la distance entre l'entrée et la sortie de l'auto-encodeur selon plusieurs méthodes et combine les résultats en fonction du problème à résoudre (procédé confidentiel).

Le score ainsi calculé est compris entre 0 et 1. Selon que le score est inférieur à 0.1 ou supérieur à 0.5, l'évènement associé est considéré normal ou incongru, les valeurs intermédiaires caractérisant les signaux faibles et les comportements peu fréquents.

## 5.8 Phase de corrélation

Notre solution, une fois entrainée à partir des données d'entrainement, est en mesure de détecter les comportements anormaux associés au scénario d'attaque, lesquels sont dilués dans les comportements normaux. Cette détection est réalisée par notre phase de corrélation déjà présentée.

## 5.9 Résultat

Notre phase de corrélation produit un graphe d'intérêt majeur représenté en **Fig. 4**.

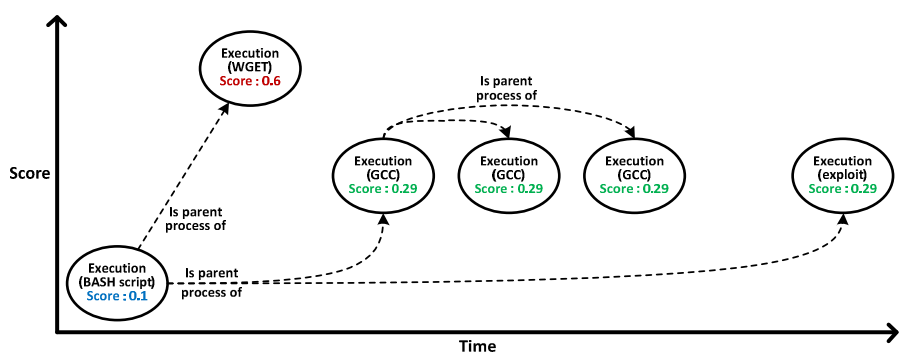

**Fig. 4.** Graphe d'intérêt majeur produit par l'IA de corrélation (données réelles).

Ce graphe comprend :

— Un signal normal résultant de l'exécution du script *bash* (score 0.1) ;
— Un signal fort correspondant à l'exécution de la commande *wget* (score 0.6) ;
— Trois signaux faibles inhérents à l'exécution de *gcc* (score 0.29) ;
— Un signal faible correspondant à l'exploit (score 0.29).

Il convient de noter que le signal associé à l'exécution du script *bash*, bien que faiblement scoré, contribue de manière significative au graphe d'intérêt majeur détecté car ce signal permet de relier l'événement *wget* aux évènements *gcc* et à l'exploit.

Le procédé génère également des faux positifs, mais ils sont liés à des tâches rarement effectuées. Il serait possible d'éliminer ces faux positifs en limitant l'importance à l'entraînement des événements très présents car les tâches de fond constituent la majeure partie des événements.

## 6 Conclusion

Notre approche associant l'apprentissage profond avec une phase de corrélation vient enrichir le modèle UEBA, montre des améliorations encourageantes en ce qui concerne le contournement des biais actuels identifiés, et produit des résultats qui revêtent un caractère explicable.

Nos premiers résultats obtenus à partir de données synthétiques issues de la simulation d'une activité sur un SI, ont été confirmés à partir de données réelles issues d'une activité sur un poste de travail.

Nos algorithmes possèdent une capacité de mise à l'échelle horizontale. En outre, nous les avons intégrés dans la suite *Elastic Stack* open-source largement répandue, afin de nous rapprocher des problématiques d'intégration avec les systèmes existants.

Nos prochains travaux adresseront les limites qui peuvent nous être opposées, d'une part en confrontant notre approche à d'autres scénarii plus réalistes et très volumineux issus de nos SOC et du produit CyberRange, d'autre part en concevant une IA de mémorisation dont l'importance nous semble cruciale.

Enfin, les présents travaux se poursuivent par une thèse en convention CIFRE.

# 7 Références